\documentclass{article}

\usepackage{PRIMEarxiv}

\usepackage[utf8]{inputenc} 
\usepackage[T1]{fontenc}    
\usepackage{hyperref}       
\usepackage{url}            
\usepackage{booktabs}       
\usepackage{amsfonts}       
\usepackage{nicefrac}       
\usepackage{microtype}      
\usepackage{lipsum}
\usepackage{fancyhdr}       
\usepackage{graphicx}       
\usepackage{algorithm}
\usepackage{algorithmic}
\usepackage{textcomp}
\usepackage{xcolor}
\usepackage{lipsum}
\usepackage{subfigure}
\title{The Morphology-Control Trade-Off: Insights into Soft Robotic Efficiency
\thanks{Corresponding author: yx388@cam.ac.uk},
\thanks{This project has received funding from the European Union’s Horizon
2020 research and innovation programme under the Marie Skłodowska-Curie grant agreement No 101034337.}
}

\author{
  Yue Xie, Kai-fung Chu \\
  University of Cambridge \\
  Cambridge\\
  UK\\
  \texttt{\{yx388,kfc35\}@cam.ac.uk} \\
   \And
  Xing Wang \\
  University of Canberra \\
  Canberra \\
  Australia\\
  \texttt{Xing.wang@canberra.edu.au} \\
  \AND
  Fumiya Iida \\
  University of Cambridge \\
  Cambridge \\
  \texttt{fi224@cam.ac.uk} \\
}

\begin{document}
\maketitle

\begin{abstract}
Soft robotics holds transformative potential for enabling adaptive and adaptable systems in dynamic environments. However, the interplay between morphological and control complexities and their collective impact on task performance remains poorly understood. Therefore, in this study, we investigate these trade-offs across tasks of differing difficulty levels using four well-used morphological complexity metrics and control complexity measured by FLOPs. We investigate how these factors jointly influence task performance by utilizing the evolutionary robot experiments. Results show that optimal performance depends on the alignment between morphology and control: simpler morphologies and lightweight controllers suffice for easier tasks, while harder tasks demand higher complexities in both dimensions. In addition, a clear trade-off between morphological and control complexities that achieve the same task performance can be observed. Moreover, we also propose a sensitivity analysis to expose the task-specific contributions of individual morphological metrics. Our study establishes a framework for investigating the relationships between morphology, control, and task performance, advancing the development of task-specific robotic designs that balance computational efficiency with adaptability. This study contributes to the practical application of soft robotics in real-world scenarios by providing actionable insights.

\end{abstract}


\section{Introduction}

Designing soft robots that achieve high task performance with minimal computational cost remains a significant challenge in robotics. Soft robots, characterized by their deformable and flexible morphologies, exhibit exceptional adaptability to unstructured dynamic environments~\cite{kim2013soft,rus2015design,shepherd2011multigait}. However, balancing the complexities of morphology and control to achieve optimal performance is still challenging. Advanced controllers can sometimes compensate for simple morphologies~\cite{cully2015robots, heess2017emergence}, but the trade-off between structural design and computational effort has deep implications for designing efficient and adaptive robotic systems. Understanding this interplay is critical for optimizing robot designs to meet the demands of real-world applications.

Morphological complexity such as structural properties such as voxel heterogeneity, symmetry, connectivity, actuator distribution~\cite{cheney2014unshackling, lipson2000automatic, sims2023evolving} influences how a robot interacts with its environment to achieve desired behaviors. Control complexity, in contrast, quantifies the computational effort required to govern these structures, often dictated by the sophistication of the control strategies employed~\cite{bengio2017deep, lecun2015deep}. These two dimensions are inherently interdependent. Despite this, much of the existing research has studied these dimensions in isolation, which limits our understanding of key questions: Can complex morphologies reduce the need for computationally demanding controllers? Conversely, when do advanced control strategies become indispensable for performance? Recent studies on co-evolution~\cite{kriegman2020scalable} highlight the need for a deeper understanding of these trade-offs, especially for scalable and robust designs.

The trade-off between morphology and control complexity is practical for resource-constrained applications, such as autonomous exploration~\cite{xin2017application}, disaster response, and wearable robotics. Robots with optimized morphologies may achieve comparable performance using simpler, energy-efficient controllers, thereby reducing power consumption and improving scalability~\cite{shepherd2011multigait, cheney2014unshackling}. Alternatively, understanding the computational limits of control strategies can guide the development of morphologies that compensate for these limitations. This trade-off lies at the core of advancing the co-design of robotic hardware and software for efficient, task-specific systems. However, current methods often fail to adequately quantify or optimize the interplay between these complexities, underscoring the need for a systematic and integrated approach.

This paper addresses these gaps by investigating the relationship between morphological complexity, control complexity, and task performance in soft robots. We propose an integrated framework that introduces practical and computationally grounded metrics to analyze these dimensions. For morphological complexity, we develop metrics that capture voxel heterogeneity, structural connectivity, and symmetry, offering a nuanced characterization of robot designs. For control complexity, we adopt Floating Point Operations per Second (FLOPs) as a scalable, architecture-agnostic metric~\cite{ptflops}, which quantifies the computational effort of executing control strategies of the reinforcement learning (RL) algorithms PPO~\cite{schulman2017proximal}.

This work contributes threefold. First, we present an integrated framework that combines morphological and control complexity metrics with RL-based optimization to analyze their interplay comprehensively. Second, we introduce interpretable and practical metrics, including FLOPs, to quantify control complexity, enabling a robust comparison of RL controllers regarding computational effort. Third, we provide valuable insights into the trade-offs between structural and computational resources, demonstrating scenarios where complex morphologies can reduce the need for sophisticated controllers and vice versa.

Through extensive experiments, we demonstrate the following:
\begin{itemize}
    \item Morphological complexity can be effectively quantified through interpretable metrics representing structural and functional attributes.
    \item The interplay between morphology and control complexity is critical for achieving high performance, with specific trade-offs emerging based on task difficulty. Low-complexity morphologies and controllers suffice for simpler tasks, whereas harder tasks require increased complexity in both domains.
\end{itemize}

\section{Related Work}
\label{sec:related}

Soft robotics, characterized by compliance and adaptability, holds transformative potential for enabling robots to interact safely and effectively with dynamic environments and humans. However, the non-linear dynamics and infinite degrees of freedom inherent to soft robots pose significant challenges for traditional control methods~\cite{trivedi2008soft}. RL has emerged as an alternative, capable of learning control policies that exploit a robot’s morphology to achieve efficient behaviors. For instance, RL has been successfully applied to complex tasks such as locomotion and object manipulation in soft robots~\cite{polygerinos2017soft}.

Evolutionary computational techniques have further advanced the co-design of soft robot morphologies and controllers. Cheney et al.~\cite{cheney2014unshackling} demonstrated that voxel-based morphologies could simplify control strategies. However, their work focused mainly on task-specific objectives, showing limited insights into broader morphology-control trade-offs. Frameworks like MAP-Elites~\cite{cully2015robots, mouret2015illuminating} have illuminated design spaces but rarely consider the interplay between structural and computational complexities. Recent studies incorporating multi-objective optimization~\cite{kriegman2020scalable} have made strides, but a systematic understanding of the relationship between morphological and control complexities remains underexplored.

The EvoGym simulation platform~\cite{bhatia2021evolution} provides a robust environment for investigating the relationship between structure and controller. Designed specifically for voxel-based soft robots, EvoGym supports customizable morphologies, diverse tasks, and efficient simulations, making it ideal for evolutionary studies. A key feature of EvoGym is its grid-based voxel representation, which allows for intricate structural designs. Each voxel represents a material type, such as soft blocks, rigid blocks, or actuators. This genome-based encoding enables systematic exploration of morphology and control, and EvoGym uses PPO~\cite{schulman2017proximal} as its default RL algorithm.

\begin{figure}[h]
\centering
\includegraphics[width=0.85\linewidth]{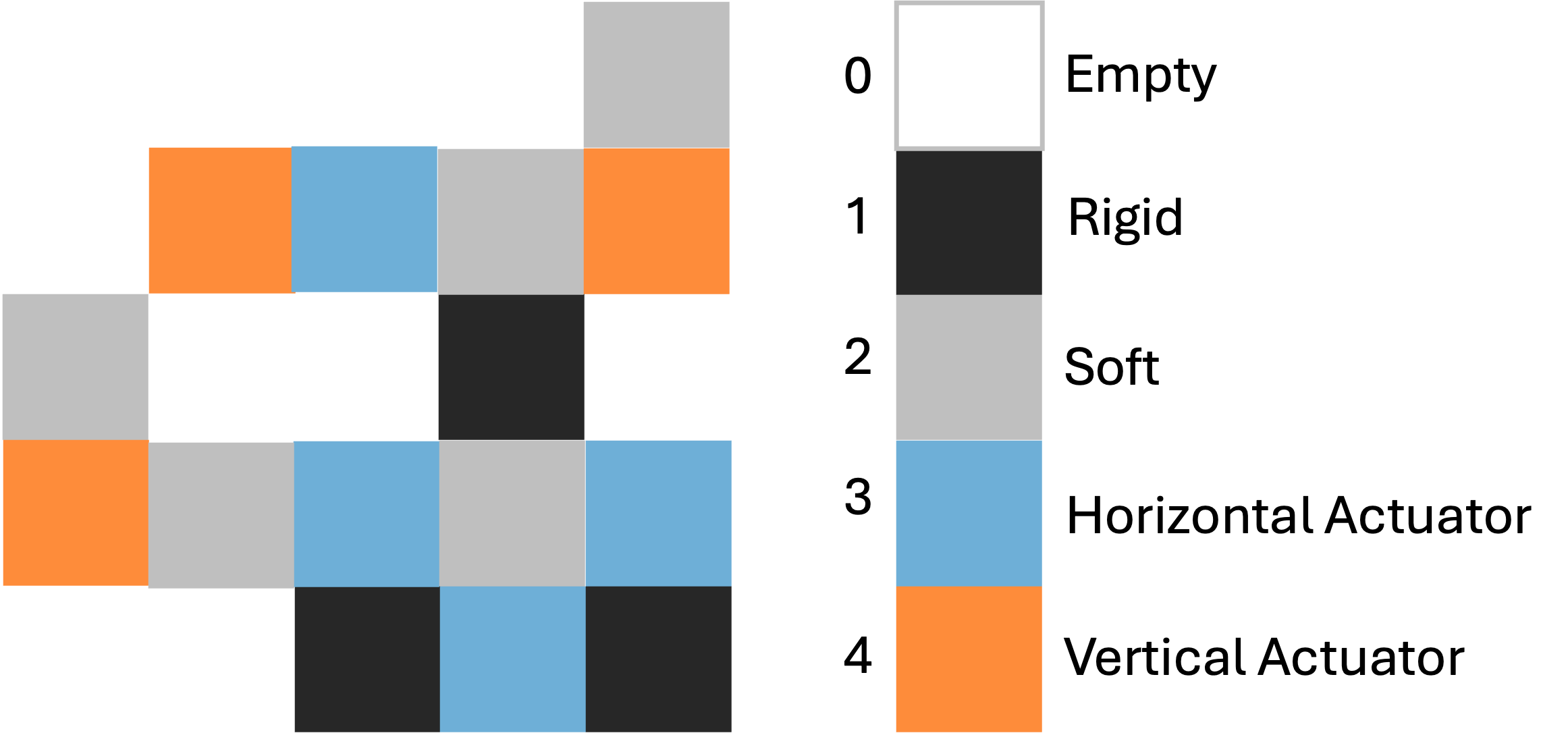}
\caption{An example of a soft robot with a $(5 \times 5)$ grid size in EvoGym.}
\label{fig:vsr}
\end{figure}

The structure of a robot in EvoGym is encoded as a genome—a grid-based matrix where each entry corresponds to a specific voxel type. Figure~\ref{fig:vsr} illustrates an example of a robot design with a $(5 \times 5)$ grid. This genome-based encoding provides a straightforward yet powerful method for representing a wide range of soft robot designs.

\section{Methodology}
\label{sec:method}

This study investigates the relationship between morphological and control complexities in soft robots by systematically evaluating their impact on task performance. We leverage an evolutionary optimization framework integrated with complexity metrics and RL algorithm to achieve this. This section details the approach to designing, evaluating, and optimizing soft robot morphologies and their controllers.

\subsection{Morphological Complexity}

Morphological complexity encompasses the physical properties of a robot’s structure, including its material composition, connectivity, and symmetry. This study quantifies morphological complexity through four metrics: voxel heterogeneity, structural connectivity, symmetry analysis, and actuator distribution as described in follow.

\subsubsection{Voxel Heteogeneity}

Voxel heterogeneity measures the diversity of materials within a robot’s structure. Robots with diverse voxel types often exhibit increased adaptability to dynamic tasks. Heterogeneity is quantified using normalized entropy:
\begin{equation}
    H = -\sum_{i=1}^n p_i \log(p_i),
\end{equation}
where $p_i$ represents the proportion of voxels of type $i$, and $n$ is the total number of voxel types. Higher heterogeneity reflects greater material diversity, which can enhance adaptability in dynamic tasks.

\subsubsection{Structure Connectivity}

Structural connectivity evaluates how well-connected the voxels are within the robot. Connectivity is measured using graph-based methods, where voxels are nodes and connections are edges:
\begin{equation}
    M_C = \frac{1}{|V|} \sum_{v\in V} degree(v),
\end{equation}
where $degree(v)$ represents the number of edges (connections) that a voxel $v$ has with other voxels. High connectivity can enhance the robot's robustness and stability, ensuring that the structure can withstand various stresses and maintain its integrity during movement. More formally, a higher connectivity score indicates that each voxel is well-connected to its neighbors, which can prevent parts of the robot from becoming isolated or detached under stress.

\subsubsection{Symmetry Analysis}

Symmetry is assessed by comparing the robot’s structure across its reflection axes. A symmetry score is calculated as:
\begin{equation}
    M_S = \frac{1}{N}\sum_{i=1}^N \left( 1-\frac{|V_{ij} - V_{ji}|}{V_{max}} \right),
\end{equation}
where $V_i$ and $V_{i'}$ are voxel properties at symmetric positions, $V_{max}$ is the maximum voxel value, and $N$ is the total number of voxels. 

\subsubsection{Actuator Distribution}

Actuator Distribution evaluates the spatial distribution of actuator voxels within the robot’s structure, offering insights into how evenly actuation capabilities are spread across the morphology. Let $\mathcal{A}$ denote the set of all actuator voxels, and a represent the position of an actuator voxel. The standard deviation of actuator positions is calculated as:
\begin{equation}
    M_{AD} = \sqrt{\frac{1}{|\mathcal{A}|}\sum_{a\in \mathcal{A}} (a-\mu)^2},
\end{equation}
where $\mu$ is the mean position of the actuators. A lower value of $M_{AD}$ indicates a more uniform distribution of actuators, suggesting a well-balanced design. Such uniformity can be crucial for achieving coordinated movements and maintaining stability during complex tasks.

\subsection{Control Complexity}

In evogym, a soft robot is controlled by a model-free controller trained using PPO. There are several complexity metrics on neural network parameters, such as network size or depth, which could be used as proxies for complexity. However, evaluating such parameters does not truly reflect complexity since different algorithms process information differently, and it does not reflect the controller’s real-time processing demands. 

Another approach to quantify complexity is to measure the energy consumed by the algorithm. Floating Point Operations Per Second (FLOPs) is a practical and architecture-agnostic metric for measuring instructions per second and computer performance in computing. FLOPs measure the number of floating-point operations required, which makes it particularly relevant for robotics, where computational efficiency and real-time performance are critical. Such a measure of number of floating-point operations required in a single forward pass through the controller can provide a direct and interpretable representation of the computational effort and energy consumption in the neural network as the theoretical neural network complexity metric~\cite{molchanov2017pruning} involved in executing control strategies. Hence, this paper quantifies the computational complexity of a robot’s controller using FLOPs.

The FLOPs of a controller is determined based on its architecture and the processing required for a single set of sensory inputs to generate actuation outputs. This involves analyzing the computational operations performed by each controller layer during inference. The calculation considers factors such as the number of neurons in the network layers, their connections, and the operations required for activation functions and weight multiplications. FLOPS is calculated for each controller to compare their computational demands and assess the relationship between control complexity and task performance. By examining FLOPs alongside task performance, this study explores the trade-offs between computational cost and control effectiveness. This analysis highlights the role of control complexity in achieving adaptive and high-performing behaviors in soft robots.

\begin{figure*}[t]
    \centering
    \includegraphics[width=0.85\linewidth]{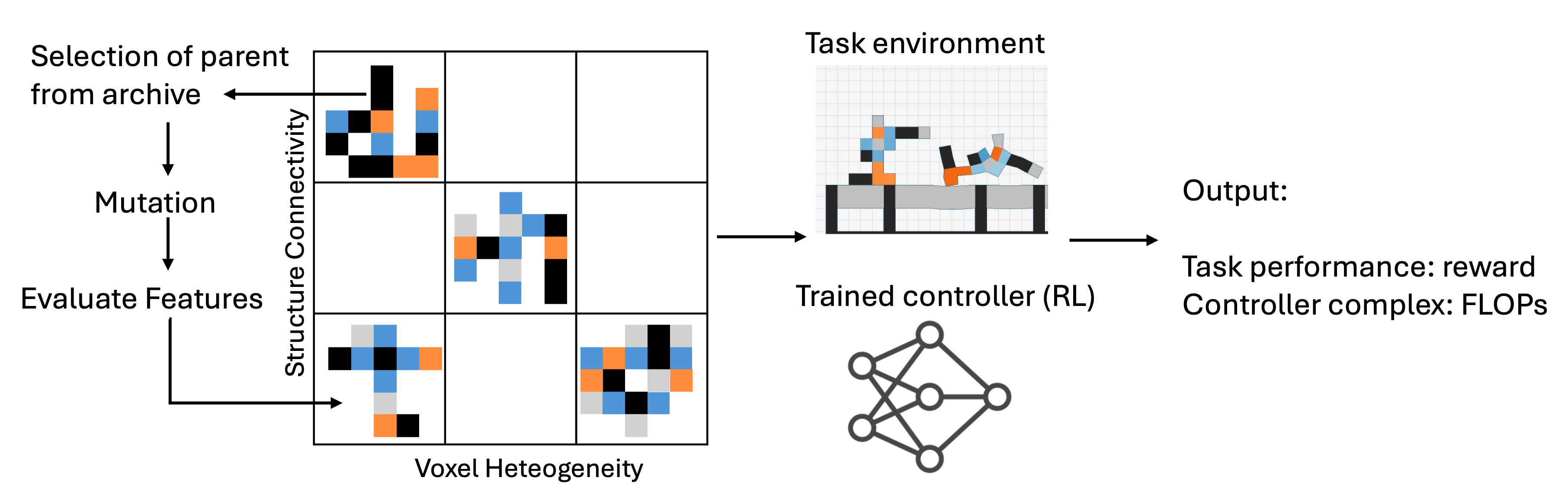}
    \caption{Exploration Framework for Morphology and Control Complexity.}
    \label{fig:flow}
\end{figure*}

\subsection{Exploration Framework}
\label{sec:explframework}

To systematically investigate the interplay between morphological complexity and control complexity in soft robots, we propose an exploration framework built upon the MAP-Elites algorithm~\cite{mouret2015illuminating}. This approach ensures the generation and evaluation of a diverse set of soft robot structures while incorporating various reinforcement learning (RL) algorithms to train controllers. The framework aims to analyze task performance and controller complexity, quantified by FLOPs, to uncover insights into their interactions.

As shown in the Figure~\ref{fig:flow}, the exploration framework begins with the generation of diverse robot structures using the MAP-Elites algorithm. MAP-Elites is a population-based evolutionary algorithm specifically designed to explore high-dimensional design spaces by promoting diversity and performance simultaneously~\cite{mouret2015illuminating}. Unlike traditional optimization methods that focus solely on finding the global optimum, MAP-Elites divides the design space into discrete bins based on predefined feature dimensions, ensuring the discovery of a wide range of solutions.

We divide the morphological design space into grids defined by four metrics of the morphological complex as introduced in the above section. Each grid corresponds to a unique combination of these metrics, representing a specific structural complexity profile. Initial robot morphologies are generated randomly and evaluated for feature metrics. The algorithm iteratively mutates selected designs, evaluates offspring, and updates the archive by filling the grids.

Algorithm~\ref{alg:diverseGener} provides the pseudocode for using MAP-Elites to generate a diverse set of robot morphologies. The process begins with initializing an archive of robot designs, calculating their feature values, and assigning them to bins based on morphological metrics. Offspring are iteratively created through mutation, evaluated for performance and features, and added to the archive if they outperform existing designs in their respective bins. This ensures a diverse and optimized set of morphologies for further analysis.

\begin{algorithm}[t]
\caption{MAP-Elites for Generating Diverse Robot Morphologies}
\begin{algorithmic}[1]
\STATE \textbf{Input} Morphological metrics $M_1, M_2, M_3, M_4$, number of iterations $N$, mutation operator.
\STATE \textbf{Output} Archive $\mathcal{A}$ 
\STATE Initialize archive $\mathcal{A}$ with empty grids based on the ranges of metrics $M_1, M_2, M_3, M_4$
\STATE Generate initial population of robot designs $\mathcal{P}$
\FOR{each design $d$ in $\mathcal{P}$}
\STATE Computer feature values $(m_1, m_2, m_3,m_4)$ for $d$
\STATE Place $d$ in the corresponding grid regarding to the feature ranges if it is empty 
\ENDFOR
\FOR{$i =1$ to $N$}
\STATE Select random design $p$ from archive $\mathcal{A}$
\STATE Generate offspring $p'$ by mutate $p$
\STATE Computer feature values $(m_1, m_2, m_3,m_4)$ for $p'$
\STATE Place $p'$ in the corresponding grid regarding to the feature ranges if it is empty 
\ENDFOR
\end{algorithmic}
\label{alg:diverseGener}
\end{algorithm}

Once the diverse set of morphologies is generated, as shown in the Figure~\ref{fig:flow}, controllers are trained using RL algorithm, and each controller provides a unique computational footprint, contributing to a range of control complexities for analysis. The controllers are trained to maximize task-specific rewards in the EvoGym simulation environment. For each controller, the computational complexity is quantified using FLOPs, which represent the number of floating-point operations required for a single forward pass through the control network.

After training, each robot-controller pair is evaluated on task-specific performance metrics. These metrics reflect the ability of the robot to perform tasks such as locomotion, obstacle avoidance, and object manipulation. Task performance is recorded as reward scores, and FLOPs are calculated to assess the computational demands of the controllers. The resulting dataset includes the morphological complexity of each robot, the task rewards and the corresponding FLOPs for each controller.

\section{Experimental Setup}
\label{sec:setup}

The experiments were conducted using EvoGym’s simulation framework, which facilitates the evolution of robot morphologies and controllers in a voxel-based environment. All evaluations were performed on a MacBook Pro with an Apple M2 chip. The computational requirements were modest, and no dedicated GPU was necessary for the experiments.

To investigate the interplay between morphological and control complexities, we employed the MAP-Elites algorithm to generate a diverse set of robot morphologies. As described in Section~\ref{sec:explframework}, this algorithm categorized robot designs into feature bins based on morphological complexity metrics, including voxel heterogeneity, structural connectivity, symmetry, and actuator distribution. The algorithm’s hyperparameters were carefully chosen to ensure a balance between diversity and task-specific performance.

For controller development, we used Proximal Policy Optimization (PPO), a state-of-the-art reinforcement learning algorithm, to train control strategies for the generated morphologies. Identical hyperparameters were applied across all experiments to ensure consistency and comparability of results, allowing us to isolate the effects of morphological and control complexities on task performance.

The reward function for each task was designed to align with task-specific objectives, with higher rewards reflecting better performance. Controllers were trained alongside morphologies during the optimization process, with fitness scores integrating both morphology and controller effectiveness. This unified training framework allowed us to analyze the complex interdependencies between morphology, control complexity, and task performance.

\begin{table}[t]
\centering
\caption{Hyperparameters for Reinforcement Learning Framework (PPO)}
\label{tab:hyperparameters}
\begin{tabular}{|l|c|}
\hline
\textbf{Parameter}                & \textbf{Value}                \\ \hline
Learning rate                     & $2.5 \times 10^{-4}$        \\ \hline
Number of steps per update        & 128                         \\ \hline
Batch size                        & 4                           \\ \hline
Number of epochs                  & 4                           \\ \hline
Discount factor ($\gamma$)        & 0.99                        \\ \hline
Entropy coefficient               & 0.01                        \\ \hline
Value function coefficient (VF Coef) & 0.5                      \\ \hline
Clip range                        & 0.1                         \\ \hline
Gradient clipping (max norm)      & 0.5                         \\ \hline
Actor-critic network architecture & 2 layers (256 units each)   \\ \hline
Optimizer                         & Adam                        \\ \hline
Total timesteps                   & $1 \times 10^{6}$           \\ \hline
Log interval                      & 50                          \\ \hline
Number of parallel environments   & 1                           \\ \hline
Number of evaluation environments & 1                           \\ \hline
Number of evaluations per interval & 1                          \\ \hline
Evaluation interval               & $1 \times 10^{5}$           \\ \hline
\end{tabular}
\end{table}

\subsection{Hyper Parameters}

The MAP-Elites algorithm was employed to generate a diverse set of robot morphologies. The algorithm utilized a population size of 100 and iterated over 1,000 generations. Morphological diversity was maintained by categorizing designs into a feature space defined by four metrics: voxel heterogeneity, structural connectivity, symmetry, and actuator distribution. Each feature was discretized into three levels, resulting in a total of 81 unique structures under evaluation.

For controllers, the hyperparameters used for training are summarized in Table~\ref{tab:hyperparameters}. These parameters were kept consistent across all experiments to ensure a fair evaluation of computational and task performance.

This experimental setup ensured that the trained controllers were assessed under uniform conditions, enabling a reliable analysis of the trade-offs between morphological complexity, control complexity, and task performance.

\subsection{Tasks}
We employed three tasks for the experiments, outlined as follows. 
\begin{itemize}
    \item Walker-v0: Easy-difficulty task of locomotion, with reward based on distance moved.
    \item BidirectionalWalker-v0: Medium-difficulty task of changing movement direction towards a randomly switching goal point, with reward based on distance moved towards the goal.
    \item ObstacleTraverser-v1: High-difficulty task of traversing an uneven path, with fitness determined by total horizontal distance moved. 
\end{itemize}

\section{Results and Analysis}
\label{sec:result}

\begin{figure*}[th]
    \centering
    \subfigure{
    \begin{minipage}{0.3\linewidth}
        \centering
    \includegraphics[width=\linewidth]{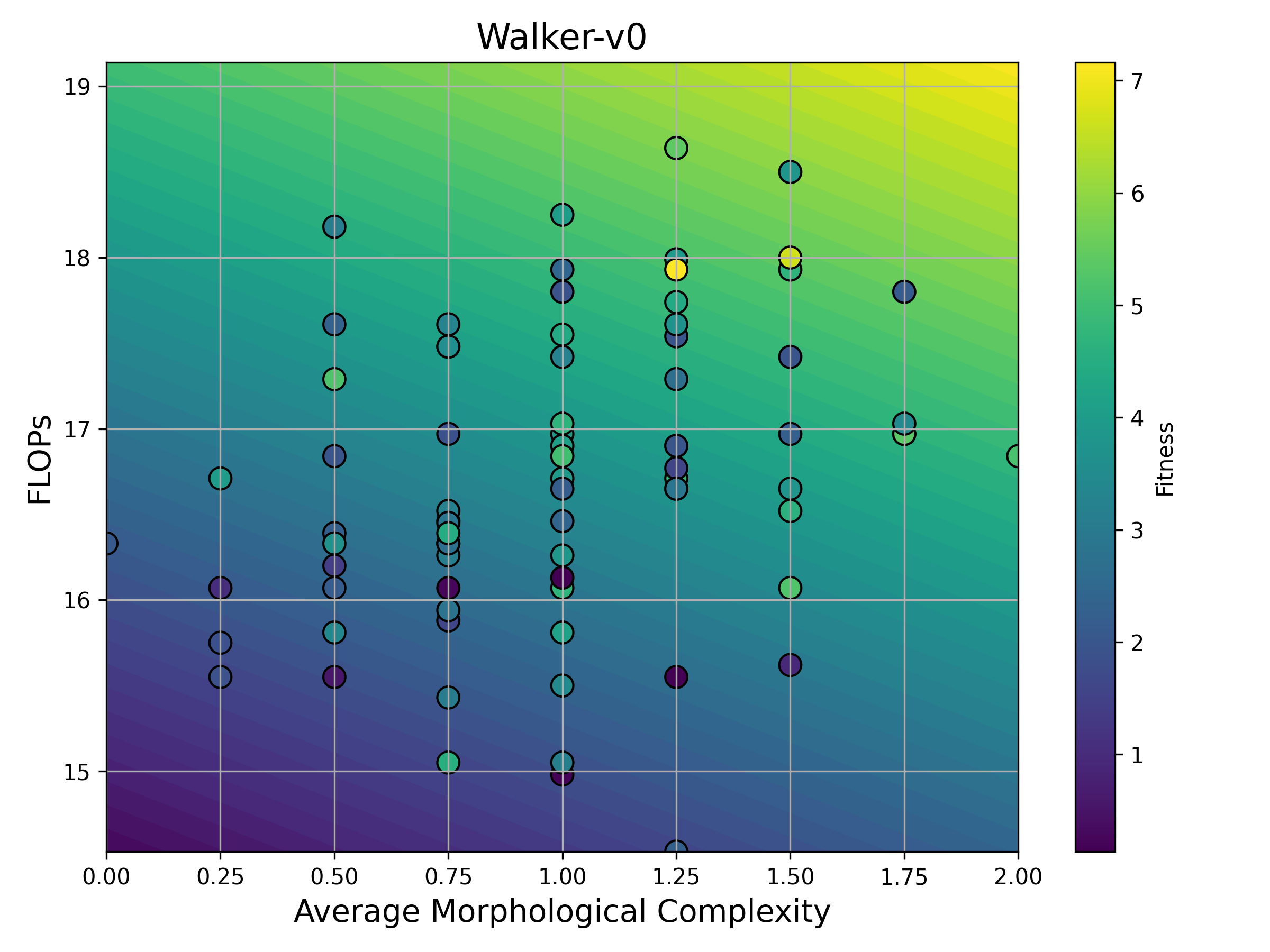}
    \end{minipage}
    }
    \subfigure{
    \begin{minipage}{0.3\linewidth}
        \centering
    \includegraphics[width=\linewidth]{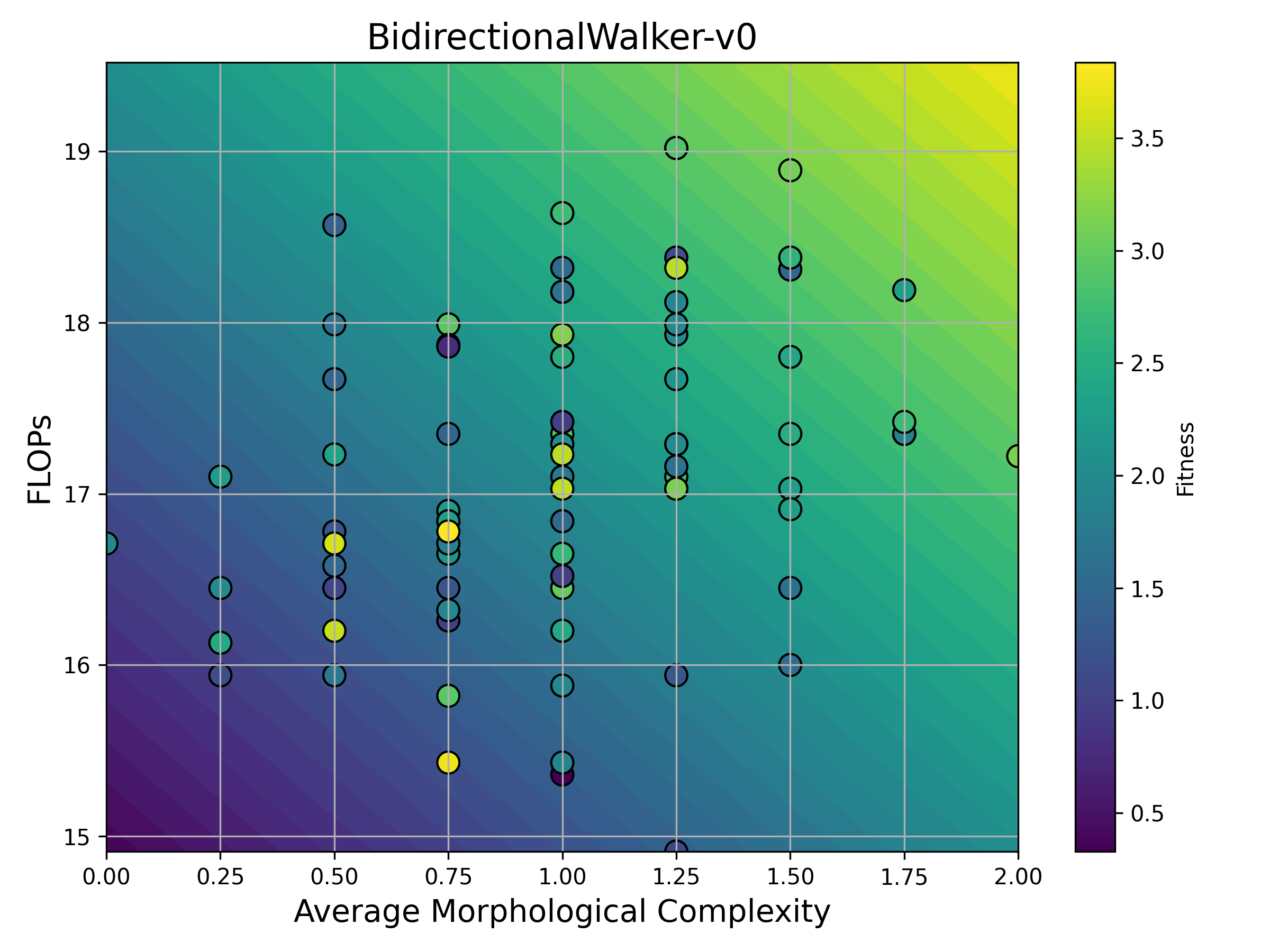}
    \end{minipage}
    }
    \subfigure{
    \begin{minipage}{0.3\linewidth}
        \centering
    \includegraphics[width=\linewidth]{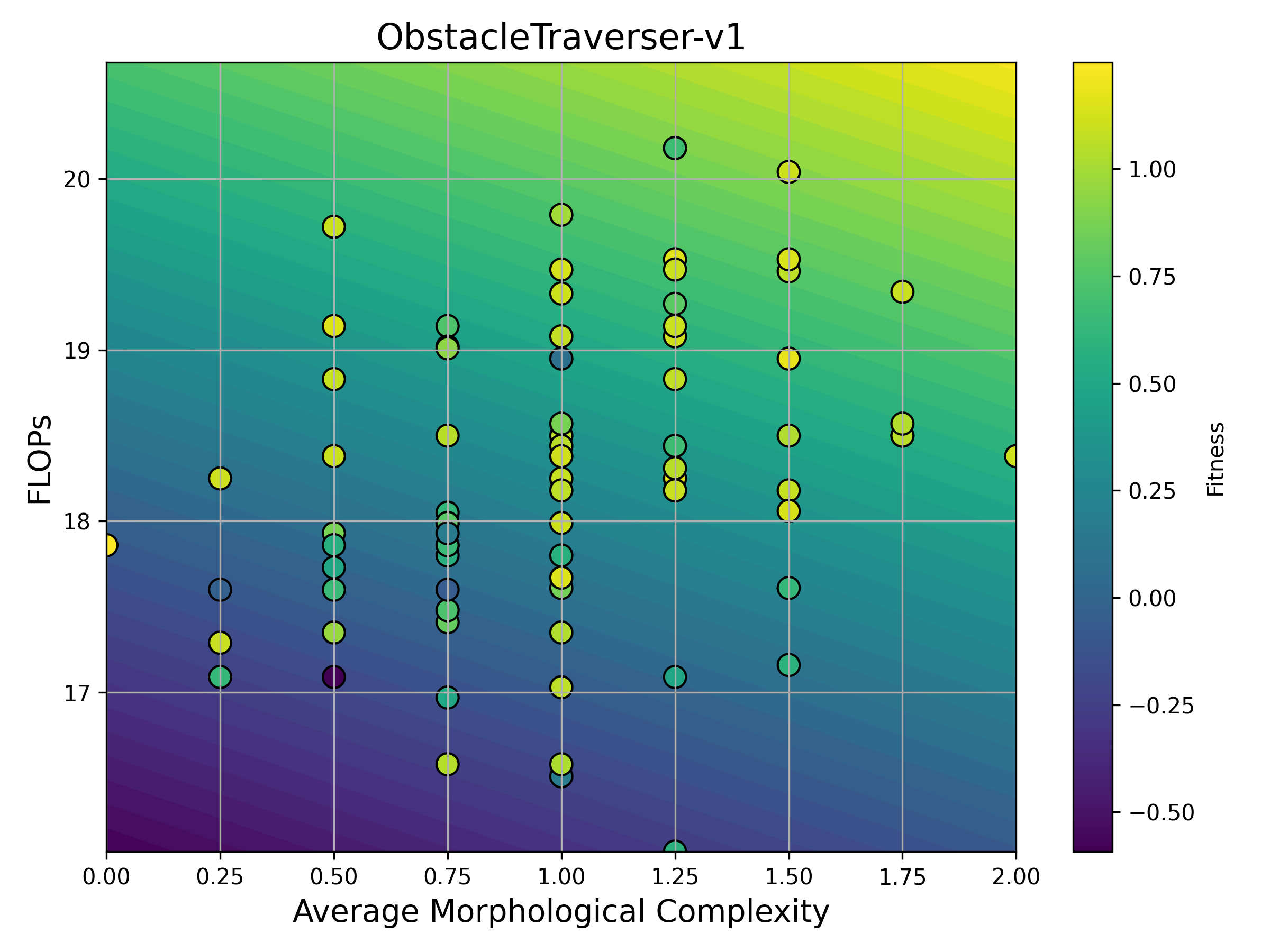}
    \end{minipage}
    }
\caption{Scatter plots illustrating the trade-off between average morphological complexity and control complexity (FLOPs) across three tasks using the PPO controller.}
\label{fig:dot_ppo}
\end{figure*}

\subsection{Trade-Off Between Morphological Complexity and Control Complexity}

\begin{table}[th]
    \centering
     \caption{Regression coefficients and $R^2$ values for each task}
    \begin{tabular}{|l|l|l|l|l|}
    \hline
       Task & $\beta_0$   & $\beta_1$ & $\beta_2$  & $R^2$\\
       \hline
       Walker & -7.169022 & 0.605433 & 0.581487  & 0.193674 \\
       \hline
       BidirectionalWalker & 0.512562 & 0.194687 & 0.084043  & 0.026908 \\
       \hline
       ObstacleTraverser & -1.829218  & 0.127625 & 0.141235  & 0.194629 \\
       \hline
    \end{tabular}
    \label{tab:regression}
\end{table}

To illustrate the relationship between morphological complexity, control complexity, and task performance, we present Figure~\ref{fig:dot_ppo}, which visualizes the trade-offs across the investigated tasks using PPO as the controller for the robots. Each subplot corresponds to one task, with the x-axis representing the average morphological complexity of the robot structures, the y-axis indicating the control complexity in terms of FLOPs, and the color of each point reflecting the task fitness. The average morphological complexity for each structure is calculated by normalizing and averaging four key metrics: voxel heterogeneity, structural connectivity, symmetry analysis, and actuator distribution. Normalization ensures comparability between metrics, and their average provides a composite measure of morphological complexity. Control complexity is quantified by the FLOPs required to execute the PPO controller, offering a clear measure of the computational cost of control. Fitness, depicted by the color intensity of each point, evaluates the robot’s task performance, with warmer colors (e.g., yellow) corresponding to higher fitness values.

The scatter plots highlight distinct task-specific patterns in how morphological and control complexities impact fitness. For Walker-v0, structures with moderate morphological complexity (approximately 1.0–1.5) and higher control complexity (FLOPs > 17) generally achieve better fitness. Lower computational controllers (FLOPs < 17) tend to correspond to lower fitness, suggesting that a minimum level of control complexity is necessary for optimal performance. In contrast, the BidirectionalWalker-v0 task shows fitness peaking for structures with higher morphological complexity ( $> 1.5$), but controller complexity exhibits greater variability, with high FLOPs values not guaranteeing better performance. For ObstacleTraverser-v1, fitness is more strongly influenced by control complexity, with structures requiring higher FLOPs ( $> 18$) consistently achieving better fitness, while morphological complexity shows a weaker effect overall.

Moreover, in Figure~\ref{fig:dot_ppo}, the data samples collected during the experiments are utilised to perform a multiple linear regression analysis, fitting a hyperplane that represents the best linear approximation of the relationship between fitness and the two complexities. The coloured hyperplane serves as a reconstruction of the fitness function with respect to morphological and control complexities, which can be expressed by the following equation:
\begin{equation}
y = \beta_0 + \beta_1 x_1 + \beta_2 x_2,
\end{equation}
where $y$ denotes the fitness, $x_1$ represents morphological complexity, $x_2$ denotes control complexity, $\beta_0$ is the constant term, $\beta_1$ is the regression coefficient for $x_1$, and $\beta_2$ is the regression coefficient for $x_2$. Table~\ref{tab:regression} lists the coefficient values and $R^2$ values for all tasks, summarizing the regression results.

From the Figure~\ref{fig:dot_ppo} and the Table~\ref{tab:regression}, several key trends emerge as follows.
Each task exhibits distinct dependencies on morphological and control complexities.
\begin{itemize}
    \item  For \textbf{Walker} task, fitness is strongly influenced by a balanced contribution from both complexities. Structures with moderate morphological complexity and high control complexity achieve the best performance. 
    \item For \textbf{BidirectionalWalker}, morphological complexity has a more dominant role, with higher morphological complexity leading to improved fitness. However, the impact of controller complexity shows greater variability, indicating that morphology alone can compensate for less sophisticated control in some cases. 
    \item  For \textbf{ObstacleTraverser}, control complexity plays the most significant role, with higher FLOPs consistently leading to better fitness. Morphological complexity shows a less pronounced but still relevant contribution.
\end{itemize}
   
The fitness patterns highlight the need for alignment between morphology and control. In Walker-v0, for example, both low-complexity morphologies and low-complexity controllers result in poor fitness, emphasizing the interdependence of these factors. Structures with optimal performance are found along regions where both morphological and control complexities are balanced. This trade-off is further supported by the diagonal fitness patterns observed in the regression hyperplanes, where the color gradient (fitness) along the diagonal lines (from top-left to bottom-right) remains consistent. This suggests that achieving fitness is not about maximizing a single complexity dimension but finding an optimal trade-off between the two.

The regression models provide valuable approximations of task fitness, but the $R^2$ values reveal that additional factors—such as environmental interactions or specific task dynamics—play significant roles in performance. For Walker-v0 and ObstacleTraverser-v1, the regression captures a notable proportion of the variance, indicating that the complexities have a direct and measurable impact on fitness. However, for BidirectionalWalker-v0, the low $R^2$ value suggests that other un-modeled factors, such as dynamic task-specific interactions, contribute significantly to fitness outcomes.

The combined scatter plots and regression hyperplanes provide a comprehensive view of how design (morphology) and computation (control) contribute to task performance, offering valuable insights for optimizing soft robot systems for specific objectives. The results and hyperplanes also demonstrate the core claim of this paper: the trade-off between morphological complexity and control complexity. The regression hyperplanes in Figure~\ref{fig:dot_ppo} visually confirm that regions of balanced trade-offs (diagonal lines with consistent colors) yield consistent fitness levels, reinforcing the hypothesis that neither extreme simplicity nor extreme complexity is ideal in isolation.

Table~\ref{tab:regression} further supports these findings by quantifying the contributions of morphological and control complexities to task performance. For tasks like ObstacleTraverser-v1, the dominance of control complexity suggests that computational advancements in controllers could significantly improve performance. Conversely, for BidirectionalWalker-v0, designing better morphologies may be a more effective approach than focusing solely on controllers.

\subsection{Sensitivity Analysis of Morphological Complexity}

\begin{figure}[th]
    \centering
    \subfigure{
    \begin{minipage}{0.95\linewidth}
        \centering
    \includegraphics[width=\linewidth]{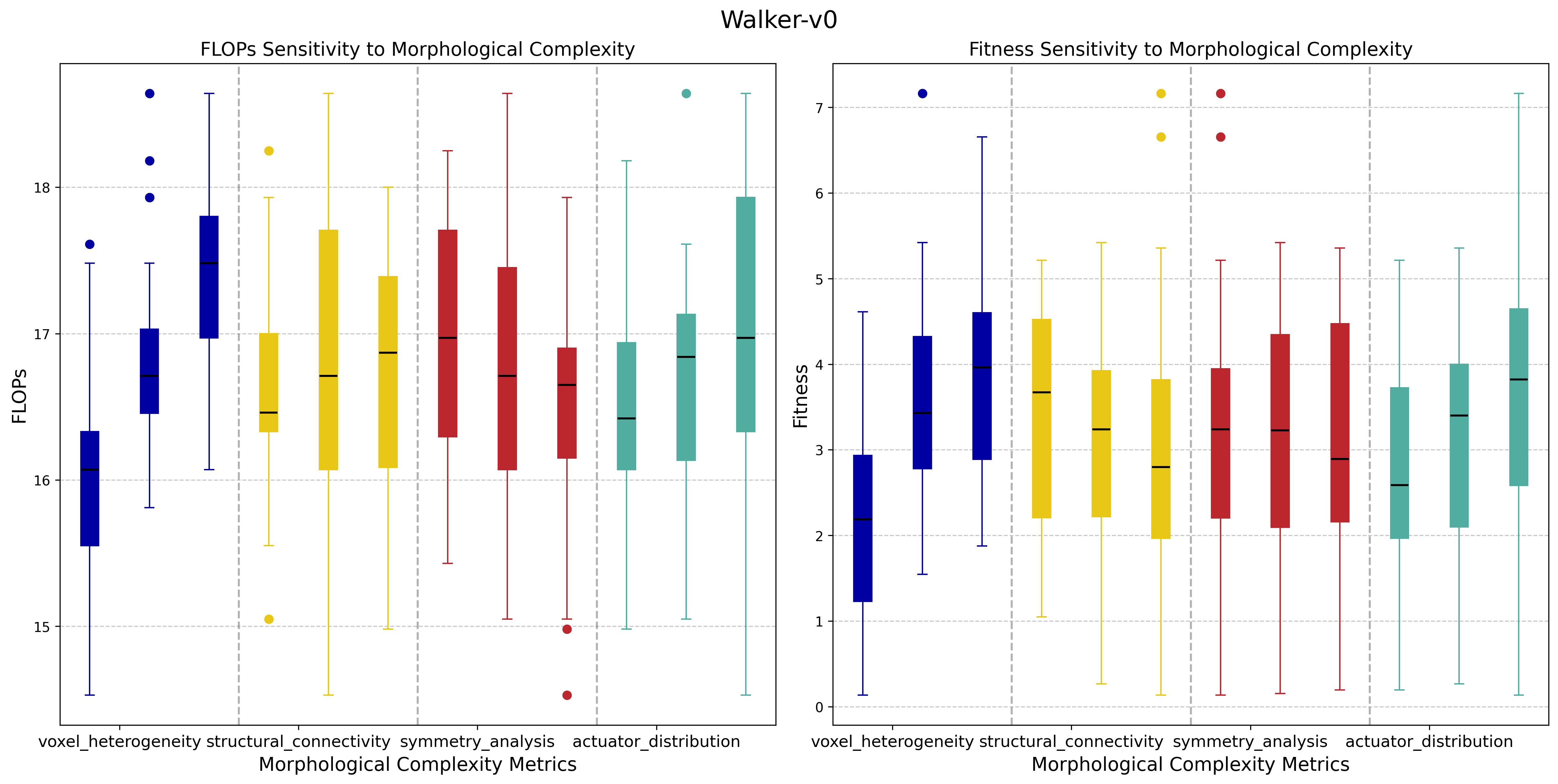}
    \end{minipage}
    }
    \subfigure{
    \begin{minipage}{0.95\linewidth}
        \centering
    \includegraphics[width=\linewidth]{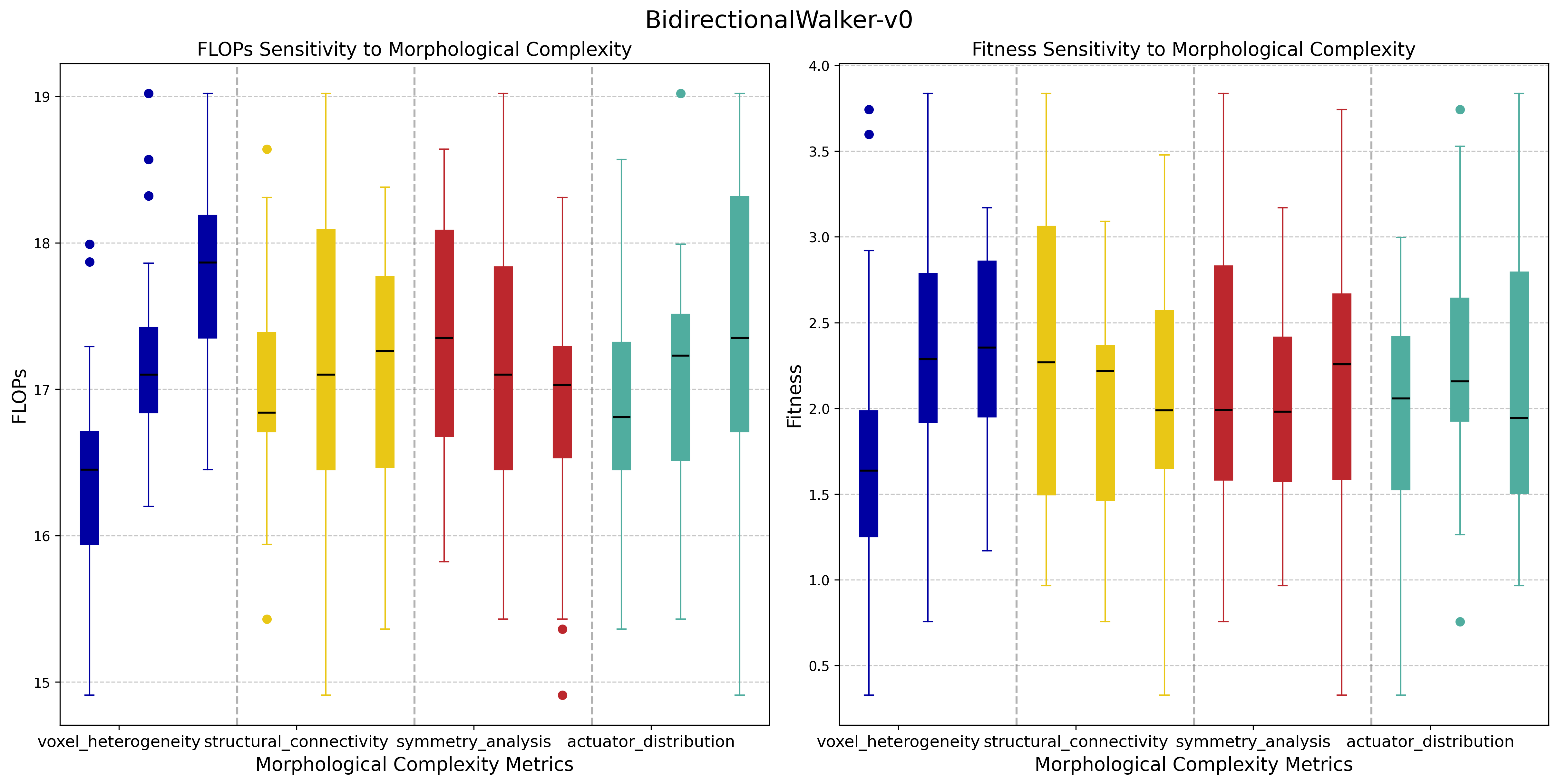}
    \end{minipage}
    }
    \subfigure{
    \begin{minipage}{0.95\linewidth}
        \centering
    \includegraphics[width=\linewidth]{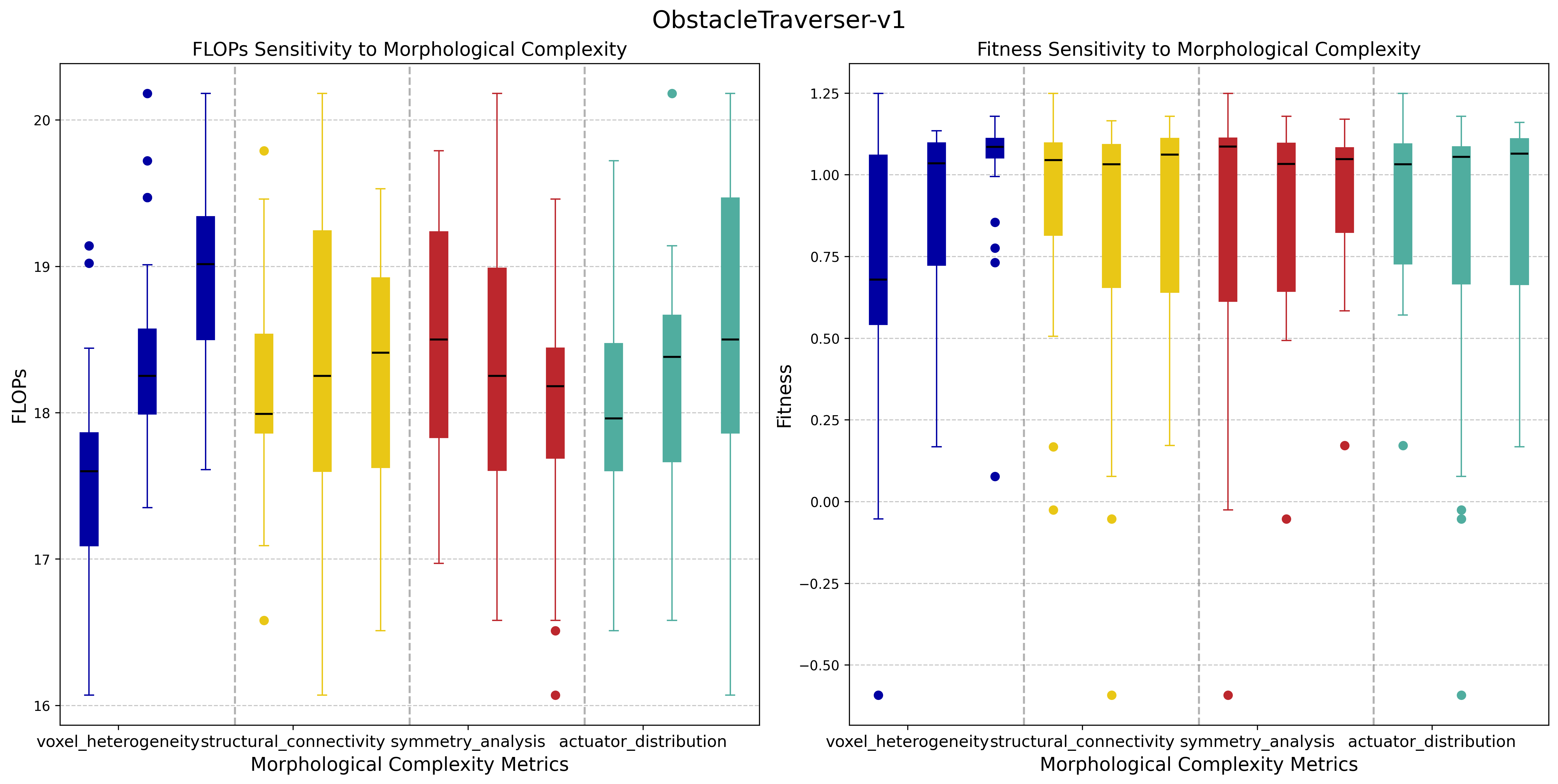}
    \end{minipage}
    }
    \caption{Sensitivity analysis of FLOPs and fitness to morphological complexity metrics across tasks. Each metric—voxel heterogeneity, structural connectivity, symmetry analysis, and actuator distribution—is represented by distinct colors. Vertical dashed lines separate the metrics for clarity, and distributions highlight variability, median, and extreme values.}
    \label{fig:boxplot}
\end{figure}

To analyze the influence of individual morphological complexity metrics on control complexity and task performance, Figure~\ref{fig:boxplot} presents box plots for all tasks. The left panels display the sensitivity of FLOPs to each complexity metric, while the right panels illustrate the relationship between fitness and the same metrics. Each morphological complexity is depicted separately with distinct colors, showing the spread, median, and extreme values of FLOPs and fitness.

As can be observed from the figure, voxel heterogeneity consistently shows a strong impact on both FLOPs and fitness across tasks. In ObstacleTraverser-v1, higher voxel heterogeneity leads to increased control complexity, with significant variability observed. Moderate voxel heterogeneity is associated with higher fitness in tasks like Walker-v0 and BidirectionalWalker-v0, indicating an optimal range where performance peaks before declining at extreme levels.

Structural connectivity exhibits a more nuanced impact, with relatively stable FLOPs and fitness distributions in Walker-v0 and greater variability in ObstacleTraverser-v1. This suggests that structural connectivity’s effect may depend on interactions with other metrics, especially in tasks with more dynamic environments. Symmetry analysis contributes differently across tasks. While fitness improves with higher symmetry in BidirectionalWalker-v0, it remains relatively stable in Walker-v0 and ObstacleTraverser-v1, highlighting task-specific responses to symmetry.

Actuator distribution demonstrates broad variability in FLOPs and fitness across all tasks, reinforcing its importance in task-specific robot designs. In BidirectionalWalker-v0, diverse actuator configurations correlate with improved fitness, emphasizing the role of actuator placement in achieving optimal performance.

The results highlight the interplay between morphological complexity metrics, control complexity, and task performance. Voxel heterogeneity and actuator distribution emerge as key drivers across tasks, while structural connectivity and symmetry have more context-dependent effects. These insights emphasize the need to tailor morphological designs to specific tasks, balancing control demands with performance outcomes.

\section{Conclusion}
\label{sec:con}

This study provides a foundational investigation into the relationship between morphological complexity, control complexity, and task performance in soft robotics. Prior research has largely examined these dimensions in isolation, leaving the interdependence between structure and control underexplored. Extensive simulation experiments conducted on Evogym, we demonstrate that performance is highly dependent on the alignment between morphology and control, with a clear trade-off emerging between these factors. Our findings show that simpler morphologies and lightweight controllers suffice for easier tasks, while more complex structures and computationally intensive controllers are essential for harder tasks. Additionally, sensitivity analysis highlights the distinct contributions of morphological complexity metrics, offering deeper insights into their role in robotic adaptation and efficiency.

While this study focuses on a single reinforcement learning controller (PPO) and a limited set of tasks, future work should extend this framework by incorporating diverse RL algorithms, such as SAC and DDPG, to evaluate their impact on control complexity and performance. Additionally, integrating performance metrics like energy efficiency and robustness could enhance the practical design of adaptive soft robotic systems.


\bibliographystyle{unsrt}  
\bibliography{references}

\end{document}